\documentclass[10pt,twocolumn,letterpaper]{article}

\usepackage{cvpr}              

\usepackage[dvipsnames]{xcolor}

\definecolor{cvprblue}{rgb}{0.21,0.49,0.74}
\usepackage[pagebackref,breaklinks,colorlinks,citecolor=cvprblue]{hyperref}
\usepackage{multirow}
\usepackage{graphicx}
\usepackage{xcolor,colortbl}

\def\paperID{7} 
\def\confName{CVPR}
\def\confYear{2024}

\title{Object Aware Egocentric Online Action Detection}

\author{%
Joungbin An \quad Yunsu Park \quad Hyolim Kang \quad Seon Joo Kim \\
Yonsei University \\
\texttt{\{joungbinan, ysp7954, hyolimkang, seonjookim\}@yonsei.ac.kr}
}

\begin{document}
\maketitle
\begin{abstract}
Advancements in egocentric video datasets like Ego4D, EPIC-Kitchens, and Ego-Exo4D have enriched the study of first-person human interactions, which is crucial for applications in augmented reality and assisted living. Despite these advancements, current Online Action Detection methods, which efficiently detect actions in streaming videos, are predominantly designed for exocentric views and thus fail to capitalize on the unique perspectives inherent to egocentric videos. To address this gap, we introduce an Object-Aware Module that integrates egocentric-specific priors into existing OAD frameworks, enhancing first-person footage interpretation. Utilizing object-specific details and temporal dynamics, our module improves scene understanding in detecting actions. Validated extensively on the Epic-Kitchens 100 dataset, our work can be seamlessly integrated into existing models with minimal overhead and bring consistent performance enhancements, marking an important step forward in adapting action detection systems to egocentric video analysis.
\end{abstract}    
\section{Introduction}
\label{sec:intro}

Recent advancements in large-scale egocentric video datasets, such as EPIC-Kitchens~\cite{epickitchens100, damen2018scaling}, Ego4D~\cite{ego4d}, and Ego-Exo4D~\cite{egoexo4d}, have revolutionized the study of human interactions and behaviors from a first-person perspective. These datasets offer a wealth of varied examples of daily activities, fostering developments in fields like activity recognition, social interaction analysis, and personal assistant technologies. Such progress promises to enhance user experiences across diverse applications.

A primary challenge in this domain is the understanding of continuous, long, fluid video streams, central for applications such as augmented reality and robotics. These applications require technologies that can efficiently parse and interpret continuous video feeds, ensuring context-aware and timely interactions. Consequently, there is an urge for advanced methodologies to meet the real-time demands of dynamic environments in understanding first-person view streaming videos.

\begin{figure}[t]
    \centering
    \includegraphics[width=1\linewidth]{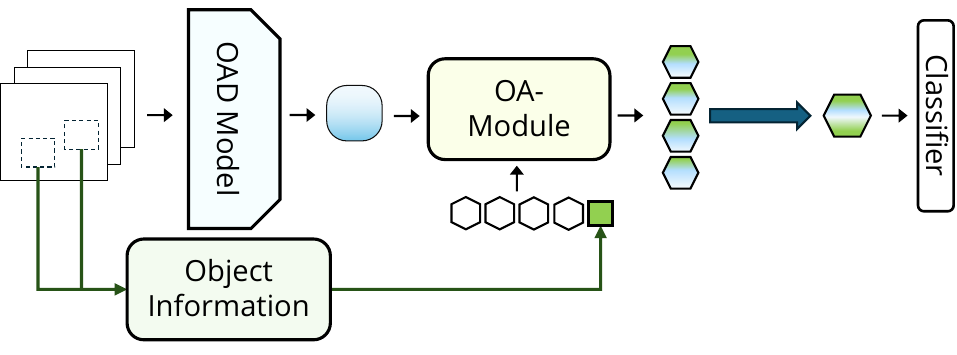}
    \caption{
    Overview of our methodology.
    }
    \label{fig:teaser}
\end{figure}

There has been ongoing work in Online Action Detection (OAD)~\cite{oadtvseries}, which involves identifying actions in a streaming video without access to future frames. However, existing works~\cite{oadtvseries, lstr, zhao2022testra, an2023miniroad} mainly focus on third-person view exo-centric videos, as its development has progressed using third-person view datasets such as THUMOS~\cite{idrees2017thumos}, TVSeries~\cite{oadtvseries}, and FineAction~\cite{liu2022fineaction}. With the recent developments in egocentric datasets, recent works~\cite{egogoalstep, egoexo4d} have released benchmarks for understanding key steps in streaming egocentric videos by employing existing OAD models. While these benchmarks have provided valuable insights, employing existing models specialized in exocentric view fails to fully exploit the unique attributes of egocentric video that set it apart from traditional third-person footage.



The main limitation stems from the inherent differences in perspective and context between exocentric and egocentric videos. Egocentric videos, characterized by their first-person viewpoint, necessitate a nuanced, object-centered understanding of the scene. This is due to the lack of pose information, a consequence of the camera’s positioning in egocentric setups. For example, in a cooking scenario, an egocentric video might capture close-up actions such as the user’s hands slicing vegetables or stirring a pot. In contrast, an exocentric video could provide a broader view of the user’s full body and the kitchen layout but fail to detail the fine-grained movements and object interactions typical of the egocentric perspective. Therefore, models designed for egocentric videos must prioritize hand-object interactions to more accurately identify specific actions like chopping, stirring, or pouring.

Recognizing this need, we instantiate an effort to integrate egocentric-specific priors into existing methodologies, thereby boosting its effectiveness in interpreting the first-person footage. 
We specifically focus on the fact that the perceived objects present in the scene offer a rich contextual background for understanding fine-grained activity. For instance, an apple placed on top of the cutting board with a knife immediately signals the impending action of `\textit{Cutting Apple}.' Similarly, the interaction patterns observed with objects commonly found in certain environments---such as utensils, appliances, and ingredients---can provide important cues for recognizing specific activities. 

As such, we aim to enhance scene object awareness in prior models by introducing an Object-Aware Module. This module begins by integrating an off-the-shelf detector~\cite{fasterrcnn} with the existing OAD framework. It utilizes learnable vectors to attend to the detected objects' outputs, extracting object scene information. These vectors are further refined by incorporating temporal cues from the OAD model, creating enriched, object and temporally aware representations. These can then be processed to determine the impending action. The overview of our methodology is shown in Fig. \ref{fig:teaser}. 


We validate our methodology through extensive experiments on the Epic-Kitchens-100 dataset using the most recent state-of-the-art OAD models. Due to its lightweight and versatile design, our Object-Aware Module can be seamlessly integrated into any existing OAD works with minimal overhead, notably enhancing the interpretation of egocentric data. Our results demonstrate consistent performance improvements across various models, confirming the module’s broad applicability and effectiveness.



\section{Related Works}
\label{sec:related_works}

\textbf{Online Action Detection} has seen considerable advancements since its introduction~\cite{oadtvseries}. Recent developments primarily fall into two categories: traditional RNN variants ~\cite{oadtvseries, trn, an2023miniroad} and transformer-based models~\cite{lstr, zhao2022testra, mat}. Notably, RNN-based approaches have recently been revisited~\cite{an2023miniroad}, demonstrating state-of-the-art performance with high efficiency. This resurgence is attributed to innovations in minimal RNN architectures and the adoption of specialized loss functions tailored for streaming video processing.

However, the majority of advancements have predominantly focused on traditional third-person view datasets. Our work diverges by extending OAD systems beyond their standard third-person constraints. We introduce a plug-and-play module that enhances object awareness, a crucial element in interpreting first-person footage. This initiative not only advances the application of OAD to egocentric videos but also lays the groundwork for future research in egocentric video analysis.

\section{Methodology}
\label{sec:method}
We present a method to augment existing OAD models in interpreting the first-person footage by making them aware of the objects in the scene. The key to this lies in extracting object information in the scene (Sec. \ref{sec:objectinformation}), adequately integrating object information and temporal cues (Sec. \ref{sec:oamodule}), and refining the updated pieces of information to detect actions more accurately (Sec. \ref{sec:classifyingactions}).

\subsection{Extracting Object Information}
\label{sec:objectinformation}
Recognizing that scene awareness of objects provides informative contextual information for understanding actions, we initiate our process by extracting object data from the scene. Drawing inspiration from notable works in the field~\cite{pirsiavash2012detecting, rulstm}, we employ Faster-RCNN~\cite{fasterrcnn}. We apply this detector to the last frame of every video snippet inputted into our model. From the output of $K$ detected bounding boxes, we aggregate their confidence scores corresponding to their category. This results in a vector $f \in \mathbb{R}^{1 \times C}$ where $C$ is the number of classes. Each value in $C$ will indicate the likeliness at which the object is present in the scene.
This object representation effectively encodes the objects present in the scene and as we validate in the experimental section, enriches scene comprehension.

\begin{figure}[t]
\centering
    \includegraphics[width=0.75\linewidth]{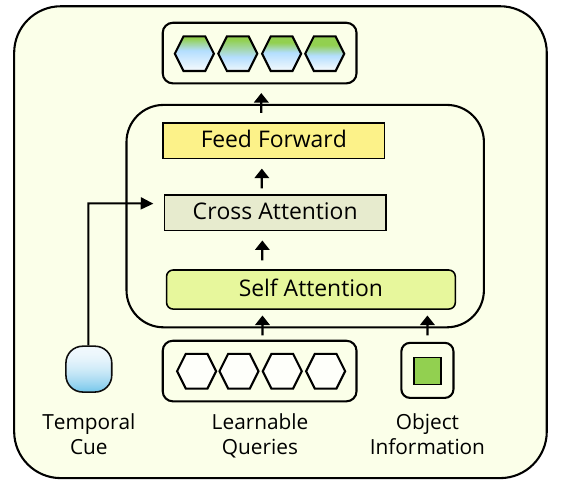}
    \vspace{-2mm}
    \caption{
    Object-Aware Module.
    }
    \label{fig:main}
    \vspace{-5mm}
\end{figure}

\subsection{Object-Aware Module}
\label{sec:oamodule}
We introduce an Object-Aware Module designed to effectively integrate object information with temporal cues. As illustrated in Fig. \ref{fig:main}, our module is composed of two transformer layers. In the first layer, learnable query vectors and object information—extracted as detailed in Sec. \ref{sec:objectinformation}—interact to update the query vectors. These vectors are then processed through the second transformer layer, where they undergo cross-attention with temporal cues captured by the OAD model. Subsequently, the updated queries pass through a feed-forward layer, preparing them for action classification. The first layer enhances the query vectors' sensitivity to scene-specific object information, fostering object awareness. The second layer employs cross-attention to merge this object information with temporal cues, crucial for accurately identifying ongoing actions. 

Despite its straightforward design and implementation, the module is effective in enhancing action identification from the first-person footage as validated in Section~\ref{sec:results}. This efficacy is attributable to the query-propagating design, which has demonstrated considerable success across various domains, including object detection~\cite{carion2020detr}, instance segmentation~\cite{cheng2022maskformer}, and action localization~\cite{tan2022pointtad}. We have adapted and refined this design for our module, enabling it to serve as an efficient information bottleneck that effectively encapsulates object information.

\subsection{Classifying Actions}
\label{sec:classifyingactions}
Finally, the updated information must be aggregated effectively to enhance action detection accuracy. The Object-Aware Module generates $N$ learnable query vectors, capturing both object and temporal data essential for action classification. A global max pooling operation then aggregates these vectors to consolidate the $N$ vectors into a single vector. This consolidated vector is subsequently classified using standard classifiers. Given that the Epic-Kitchens dataset requires the classification of Verbs, Nouns, and Actions separately, we utilize three distinct classifiers tailored to each category. 

\section{Experimental Results}
\subsection{Dataset}
We conduct our experiments using the Epic-Kitchens-100 dataset \cite{epickitchens100}, which comprises 100 hours of egocentric video featuring 90,000 action segments. These segments are annotated and categorized into 97 verb classes, 300 noun classes, and 3,806 action classes. Our experiments follow the training and validation split outlined by Furnari et al.~\cite{rulstm}.

\subsection{Experimental Setup}
\noindent \textbf{OAD Models.} We employ the open-source implementations of TeSTra~\cite{zhao2022testra}, MAT~\cite{wang2023memory}, and MiniROAD~\cite{an2023miniroad} in their default configurations. Notably, since TeSTra and MAT implement the MixCLIP data augmentation technique, we also apply MixCLIP in our use of MiniROAD. 

\noindent \textbf{Setting.} We follow the standard OAD protocol: videos are processed at 24 fps and frames are fed into a TSN model \cite{tsn} pretrained on the Kinetics-400 dataset \cite{i3dkinetics}, using a window size and stride of 6.
Regarding the module configuration, we utilize a single block of the Object-Aware Module with 16 learnable queries and an embedding dimension of 1024 within the module. We project the object information with a single feed-forward layer to match the dimension of query vectors. 

\noindent \textbf{Evaluation Metrics.} We follow the dataset setting ~\cite{epickitchens100} and use the mean Top-5 Recall to measure Verb, Noun, and Action accuracy.

\begin{table}[t]
\centering
\resizebox{\columnwidth}{!}{%
\begin{tabular}{c|ccc}
\toprule
\multirow{2}{*}{\textbf{Method}} & \multicolumn{3}{c}{\textbf{Overall}}              \\
                        & \textbf{Verb}         & \textbf{Noun}        & \textbf{Action }     \\ \hline
TeSTra~\cite{zhao2022testra}                  & 39.7         & 45.6        & 25.1        \\
MAT~\cite{mat}                     & 44.5         & 48.3        & 26.3        \\
MiniROAD~\cite{an2023miniroad}               & 39.2         & 42.7        & 22.7        \\ \hline
TeSTra + OA-M           & 50.5 (\textbf{+10.8}) & 47.6 (\textbf{+2.0}) & 25.2 (\textbf{+0.1}) \\
MAT + OA-M              & \cellcolor{lightgray}52.9 (\textbf{+8.4})  & \cellcolor{lightgray}48.7 (\textbf{+0.4}) & 26.4 (\textbf{+0.1}) \\
MiniROAD + OA-M         & 49.2 (\textbf{+10.0}) & 48.5 (\textbf{+5.8}) & \cellcolor{lightgray}26.7 (\textbf{+4.0}) \\ \bottomrule
\end{tabular}%
}
\vspace{-2mm}
\caption{Recent OAD models (top) vs. integration of our Object-Aware Module (bottom).}
\label{tab:main}
\end{table}

\begin{table}[t]
\centering
\resizebox{0.75\columnwidth}{!}{%
\begin{tabular}{c|ccc}
\hline
\textbf{Object Information} & \textbf{Verb} & \textbf{Noun} & \textbf{Action} \\ \hline
None (baseline)             & 39.2          & 42.7          & 22.7            \\
To Input (with RGB)         & 38.6          & 42.5          & 22.6            \\
OA-Module                & \textbf{49.2} & \textbf{48.5} & \textbf{26.7}   \\ \hline
\end{tabular}%
}
\vspace{-2mm}
\caption{Position to input object information.}
\label{tab:main2}
\vspace{-2mm}
\end{table}

\subsection{Results}
\label{sec:results}
\paragraph{Main Results}
The results of applying our methodology to three recent OAD models are presented in Table \ref{tab:main}. We observe consistent improvements in Verb, Noun, and Action accuracy across all models tested. These findings confirm the versatility of our Object-Aware Module, demonstrating its capability to be seamlessly integrated into any existing OAD framework. Importantly, as validated by improvements in all Verb, Noun, and Action accuracies, it enhances object awareness, which is crucial for accurately interpreting egocentric scenes. We observe the most significant improvements in Verb accuracy, as identifying the object of interaction provides important clues about the impending action, such as opening, squeezing, cutting, etc. However, gains in Action accuracy are less pronounced, which we attribute to the complex nature of the dataset’s action labels, encompassing over 3,800 distinct classes.

\vspace{-3mm}
\paragraph{How to utilize object information}
Table \ref{tab:main2} shows the outcomes from various methods of integrating object information using the state-of-the-art OAD model \cite{an2023miniroad}. The baseline, presented in the first row, does not include object information. The second row illustrates results when object information is concatenated with RGB at the input level. In contrast, the last row highlights our approach using the OA-Module. These results indicate that na\"ively adding object information does not yield optimal outcomes, emphasizing the critical role of the OA-Module in effectively leveraging this data.

\section{Limitations}
As this represents the preliminary stage of an ongoing project, our experimental results are currently somewhat limited. We acknowledge that our incorporation of object scores for providing object information, as described in Section~\ref{sec:method}, could be further improved. We are currently working on using only the ``active'' objects that the user is actually interacting with, instead of all the objects present in the scene. Also, instead of a na\"ive objectness score, we are experimenting to incorporate the active object's spatial information. We plan to extend this framework to more diverse datasets, including Ego4D and Ego-Exo4D. 

\section{Conclusion}
This work aims to enhance existing Online Action Detection frameworks for more effective application in egocentric videos by leveraging egocentric priors. We introduce the Object-Aware Module, a flexible module that can be seamlessly integrated into any current OAD method. This module utilizes object information as important contextual cues to predict impending actions better. Through this, we improve the adaptability and accuracy of OAD systems in interpreting first-person video data, paving the way for advancements in applications that require precise real-time analysis. 
{
    \small
    \bibliographystyle{ieeenat_fullname}
    \bibliography{main}
}


\end{document}